\documentclass{article}

\usepackage[final]{ml4eng_2020}

\usepackage[utf8]{inputenc} 
\usepackage[T1]{fontenc}    
\usepackage{hyperref}       
\usepackage{url}            
\usepackage{booktabs}       
\usepackage{amsfonts}       
\usepackage{nicefrac}       
\usepackage{microtype}      
\usepackage{amsmath}
\usepackage{graphicx}

\DeclareMathOperator*{\argmin}{argmin}
\usepackage{longtable}
\usepackage[ruled,vlined]{algorithm2e}
\usepackage{float}

\newcommand{\loc}[1][i]{x_{#1}}

\newcommand{\glbVar}[1]{#1}
\newcommand{\cpyVar}[1]{\bar{#1}}

\newcommand{\glb}{\glbVar z}
\newcommand{\cpy}[1][i]{{\cpyVar z}_{#1}}

\usepackage{xcolor}

\title{Collaborative Multidisciplinary Design Optimization \newline with Neural Networks}

\author{%
  Jean de Becdelièvre \\ Stanford University\\
  \texttt{jeandb@stanford.edu} \\
   \And  
    Ilan Kroo \\
   Stanford University\\
   \texttt{kroo@stanford.edu} \\
}

\begin{document}

\maketitle

\begin{abstract}
The design of complex engineering systems leads to solving very large optimization problems involving different disciplines. 
Strategies allowing disciplines to optimize in parallel by providing sub-objectives and splitting the problem into smaller parts, such as Collaborative Optimization, are promising solutions.
However, most of them have slow convergence which reduces their practical use. 
Earlier efforts to fasten convergence by learning surrogate models have not yet succeeded at sufficiently improving the competitiveness of these strategies.
This paper shows that, in the case of Collaborative Optimization, faster and more reliable convergence can be obtained by solving an interesting instance of binary classification: on top of the target label, the training data of one of the two classes contains the distance to the decision boundary and its derivative.
Leveraging this information, we propose to train a neural network with an asymmetric loss function, a structure that guarantees Lipshitz continuity, and a regularization towards respecting basic distance function properties.
The approach is demonstrated on a toy learning example, and then applied to a multidisciplinary aircraft design problem.
\end{abstract}

\section{A Bilevel Architecture for Design Optimization}

\subsection{Introduction}
\label{intro}
Design optimization of complex engineering systems such as airplanes, robots or buildings often involves a large number of variables and multiple disciplines.
For instance, an aircraft aerodynamics team must choose the wing geometry that will efficiently lift the weight of the airplane. This same weight is impacted by the structures team, who decides of the inner structure of the wing to sustain the lifting loads.
While the disciplines operate mostly independently, some of their inputs and outputs are critically coupled.

An important body of research on \emph{design architectures} \citep{Martins2013} focuses on efficient approaches for solving design problems.
Convergence speed and convenience of implementation are major priorities of the engineering companies adopting them.
Solving design problems as a large single-level optimization problem has been shown to work well when all disciplinary calculations can be centralised \citep{Tedford2010}.
However, it is often impractical due to the high number of variables and to the added complexity of combining problems of different nature (continuous and discrete variables for instance).

Bilevel architectures are specifically conceived to be convenient given the organization of companies and design groups. 
An upper \textit{system} level makes decisions about the shared variables and assigns optimization sub-objectives to each disciplinary team at the \textit{subspace} level. 
The value of local variables - such as the inner wing structure - is decided at the subspace level only, which allows the use of discipline specific optimizers and simulation tools.

Collaborative Optimization (CO)\citep{braun1997collaborative, Braun1996} is the design architecture that is closest to current systems engineering practices and is the focus of our work.
The system level chooses target values for the shared variables that minimize the objective function. The sub-objective of each subspace is to match the targets, or minimize the discrepancy, while satisfying its disciplinary constraints.
Collaborative optimization has been successfully applied to various large-scale design applications, including supersonic business jets \citep{Manning1999}, satellite constellation design \citep{budianto2004design}, internal combustion engines \citep{mcallister2003multidisciplinary}, and bridge structural design under dynamic wing and seismic constraints \citep{balling2000collaborative}.

However, the widespread adoption of CO is limited by its slow convergence speed, which often is an issue with bilevel architectures \citep{Tedford2010}. 
Mathematical issues when solving the CO system-level problem with conventional sequential quadratic programming optimizers have been documented in
\citep{Demiguel2000}, \citep{Braun1996} and \citep{Alexandrov2000}, and various adapted versions have been attempted \citet{Roth2008, Zadeh2009}.

A promising approach is to train surrogate models at predicting whether each discipline will be able to match the given target value.
Using such representations of the feasible set of each discipline, the system-level problem can choose better informed target values for the shared variables.
The use of sequentially refined quadratic surrogate models is demonstrated in \citet{Sobieski2000}, and several other studies use the mean of a Gaussian process (GP) regression \citep{tao2017enhanced}. 
While the earlier work focused on predicting the discrepancies of each discipline using regression, \citep{Jang2005} recognized that the main goal is to classify targets as feasible or infeasible, and proposes an original mix between a neural network classifier and a GP regression. 
Our paper builds upon this work, and shows that the information obtained from the subspace solution can be used to improve classification performance.

The rest of the paper is organized as follows.
Section \ref{sec:co} shows how CO reformulates design problems and presents an aircraft design example problem.
Section \ref{sec:regcls} takes a deeper dive into the structure of the data available to train a classifier, and proposes to use Lipschitz networks \citep{anil2019}. It demonstrates that they generalize better on this task and provides intuition for it.
In Section \ref{sec:alg} the example problem of Section \ref{example} is solved with our approach.
Finally, Section \ref{sec:disc} concludes, discusses the drawbacks of the current approach and suggests directions for improvements.

\section{Collaborative Optimization}
\label{sec:co}
\subsection{Mathematical Formulation}

Mathematically, the original MDO problem can be written:
\begin{align}
\label{eq:mdo}
&\underset{\loc[1:n], \glb}{\text{minimize}} &  & f(\glb) \\
& \text{subject to} & & c_i(\loc, \glb) \leq 0 & \forall i \in {1 \dots N_d} \nonumber
\end{align}

where $N_d$ is the number of disciplines, $\glb$ are the shared variables and $\loc$ the local variables of discipline $i$.
Typically, the objective $f$ is easy to compute, but evaluating the constraints $c_i$ requires time consuming disciplinary specific analysis.
With CO, we form the \emph{system}-level problem that only contains shared variables:
\begin{align}
\label{eq:sys}
&\underset{\glb}{\text{ minimize}} & \quad & f(\glb) \\
& \text{subject to} & & J_i^*(\glb) \leq 0 & \forall i \in 1 \dots N_d \nonumber
\end{align}
The $(J^*_i)_{i\in {1 \dots n} }$ are the values of the \emph{subspace}-level optimizations:
\begin{equation}
    \label{eq:sub}
    J_i^*(\glb) = \min_{\cpy \in \{ \cpy \vert \exists \loc \; c_i(\loc, \cpy) \leq 0\}} \Vert \cpy - \glb \Vert_2^2
\end{equation}
$\cpy$ is a local copy of the global variables. The subspace problem sets $\cpy$ to the feasible point closest to $\glb$.
In practice, the constraint $J_i^*(\glb) \leq 0$ is often replaced by $J_i^*(\glb) \leq \epsilon$, with $\epsilon$ a positive number.
\subsection{Simple Aircraft Marathon Design Example Problem}
\label{example}
We consider the design of an electric powered RC airplane to fly a marathon as fast as possible.
The fuselage, tail, motor and propeller are are given, and we must design the wing and pick a battery size.
There are two disciplines:
\begin{enumerate}
    \item Choosing a wing shape (span $b$ and area $S$) and computing its drag $D_{wing}$, weight $W_{wing}$ and lift $L$. Making sure $L$ is greater than the total weight $W = (W_{bat} + W_{fixed} + W_{wing})$. Also, computing the total aircraft drag $D$.
    \item Choosing the propulsion system operating conditions (voltage $U$ and RPM) such that the propeller torque $Q_p$ matches the motor torque $Q_m$, and computing the thrust $T$. Making sure $T$ is greater than the total drag $D = (D_{fixed} + D_{wing})$. Using $V$ and the electric power consumed by the motor $P_{in}$ to compute $W_{bat}$.
\end{enumerate}
In CO form, the \emph{system}-level problem only contains the drag $D$ and speed $V$:
\begin{align*}
    &\min_{\glbVar{D}, \glbVar{V}, \glbVar W_{bat}} &&  - \glbVar{V}\\ 
    & \text{subject to} & &J^*_{aerodynamics}(\glbVar{D},\glbVar{V},\glbVar W_{bat}) = 0 \\
     & && J^*_{propulsion}(\glbVar{D},\glbVar{V},\glbVar W_{bat}) = 0 
\end{align*}
where $J^*_{aerodynamics}$ and $J^*_{propulsion}$ are the optimal values of the following \emph{subspace}-problems:
\noindent
\begin{minipage}[t]{.45\textwidth}
\begin{align*}
    &\min_{
         \cpyVar D, \cpyVar V, \cpyVar W_{bat},  
          b, S 
    } 
    && 
    \begin{array}{c}
         (\cpyVar D - \glbVar{D})^2 + (\cpyVar V - \glbVar V)^2  \\
          + (\cpyVar W_{bat} - \glbVar W_{bat})^2 
    \end{array} \\ 
    & \text{subject to} && W_{wing} = \text{Weight}(b, S, L) \\
    &&& L, \cpyVar D = \text{Aero}(b, S, \cpyVar V) \\
        &&& L \geq W_{wing} + W_{fixed} + \cpyVar W_{bat}
\end{align*}
\end{minipage}%
\hfill 
\begin{minipage}[t]{0.45\textwidth}
\begin{align*}
    &\min_{
         \cpyVar D, \cpyVar V, \cpyVar W_{bat},  
          \omega, U 
    } 
    && 
    \begin{array}{c}
         (\cpyVar D - \glbVar{D})^2 + (\cpyVar V - \glbVar V)^2  \\
          + (\cpyVar W_{bat} - \glbVar W_{bat})^2 
    \end{array} \\ 
    & \text{subject to} &&Q_{m}, P_{in} = \text{Motor}(\omega, U)\\
    &&&Q_{p}, T= \text{Propeller}(\omega, \cpyVar V)\\
    &&& Q_p = Q_m , \;\; T \geq \cpyVar D \\
    &&& \cpyVar W_{bat} = \text{Battery}(\cpyVar V, P_{in})
\end{align*}
\end{minipage}

Details about the various models (Weight, Aero, Motor, Battery and Propeller), as well as a nomenclature can be found in appendix \ref{appendix:example}.
This problem is useful to explain CO, but note that it can easily be solved using single-level optimization because there is a small number of variables and each discipline is easily evaluated.

\section{From the Subspace Problem to Signed Distance Functions}
\label{sec:regcls}

\begin{figure}[t]
    \centering
    \includegraphics[width=.6\columnwidth]{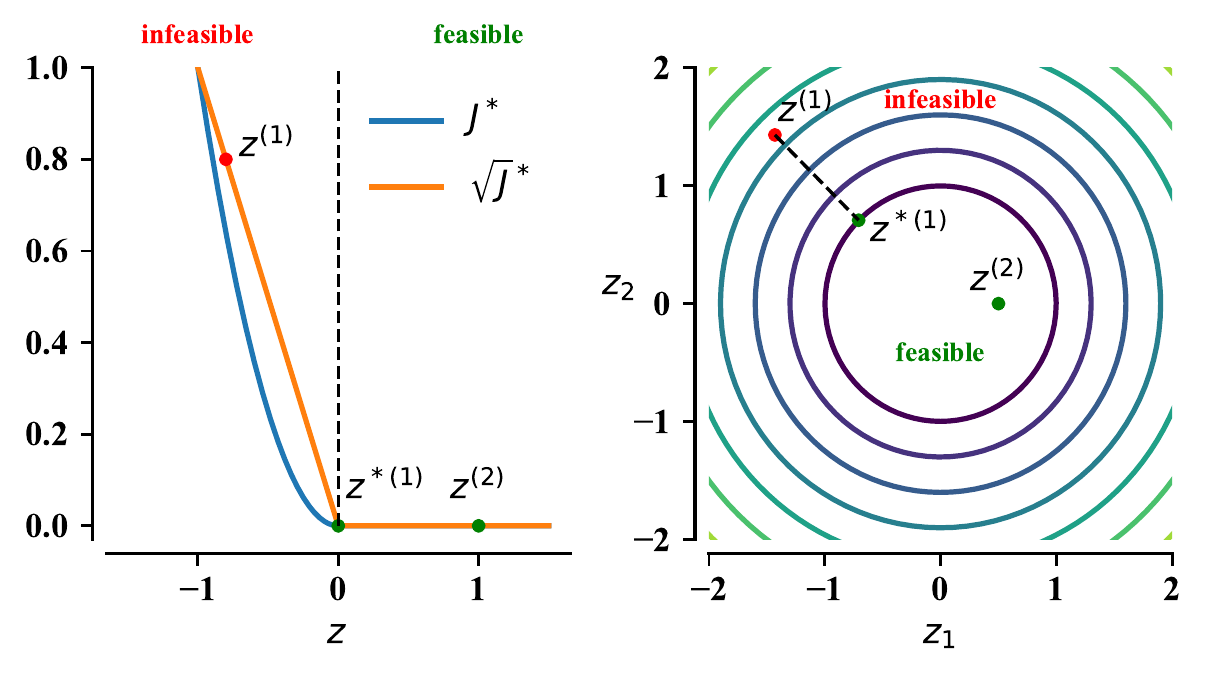}
    \caption{Plots of the square root subspace value function $\sqrt{J^*}$ for an imaginary 1-d problem where the feasible set is $\mathbf{R}^+$({left}), and an imaginary 2-d problem where the feasible set is the unit ball ({right}). The 1-d case also shows the value function $J^*$ itself, where the reader can note the null derivative at the boundary which makes classification ambiguous. In both cases, $z^{(1)}$ is shown as an example of an infeasible point (along with its feasible projection ${z^*}{(1)}$) while $z^{(2)}$ is shown as a example of a feasible point.
    }
    \label{fig:relu}
\end{figure}

\subsection{A Mixed Classification and Regression approach}
\label{sec:mix}
The subspace optimization problem (Eq. \ref{eq:sub}), is a projection operation onto the disciplinary feasible set. 
Hence, the optimal value of the $i$-th subspace $J^*_i(\glb)$ is the square of the distance between $\glb$ and closest point to $\glb$ inside the feasible set of discipline $i$: it is 0 in the feasible region and positive outside.
Figure \ref{fig:relu} shows the aspect of $\sqrt{J^*_i}$ for a 1-d and 2-d simple examples.
The figure also shows hypothetical datapoints $\glb_i$, as well as the result $\glb^*_i$ of the projection.

The main purpose of the $J^*_i(\glb) \leq 0$ constraint in the system-level problem (\ref{eq:sys}) is to indicate the feasibility of $\glb$.
As was noted in \citet{Martins2013} and \citet{Tedford2010}, this constraint is numerically ambiguous. $J^*_i(\glb) \leq \epsilon$ is typically used instead, where the choice of the threshold $\epsilon$ can negatively impact the feasibility of the final design. 
The fact the norm of the gradient of $J^*_i(\glb)$ is 0 at the boundary reinforces the ambiguity.
With surrogate models of $J^*_i$, the ambiguity becomes even more of a problem. Typical nonlinear regression approaches do not represent well the large flat areas of the feasible region and tend to oscillate around zero.

In this work we propose to train a simple feedforward neural network to classify whether a point $\glb$ is feasible or not.
While exactly regressing on $J^*_i$ does not lead to satisfying results, the information about the distance between each infeasible point and the boundary of the feasible domain remains very precious.
Let $h_i$ be a neural network and $\mathcal{D} = ({J_i^*}^{(j)}, \glb^{(j)})_{j\in {1 \dots N}}$ a set of evaluations of the $i$-th subspace problem (Eq. \ref{eq:sub}), we propose the following loss function:

\begin{equation}
    \label{eq:loss}
    l(\glb^{(j)}) = \begin{cases}
                    \vert h_i(\glb^{(j)}) - \sqrt{{J_i^*}^{(j)}} \vert & \text{if } \glb^{(j)} \text{ is infeasible}\\        
                    \max(h_i(\glb^{(j)}), 0)& \text{otherwise}
                \end{cases}
\end{equation}

The network is only equal to $\sqrt{J_i^*}$ in the infeasible region. In the feasible region, it is trained to be non-positive. The change of sign at the boundary creates a non-ambiguous classification criterion.

In practice, each subspace solution at $\glb^{(j)}$ yields a projected point ${\glb^*}^{(j)}$ and the gradient of $J^*_i(\glb^{(j)})$. This information can easily be included in the loss function in equation \ref{eq:loss}, see appendix \ref{appendix:subspace} for details.

\begin{figure}[t]
    \centering
    \includegraphics[width=.6\columnwidth]{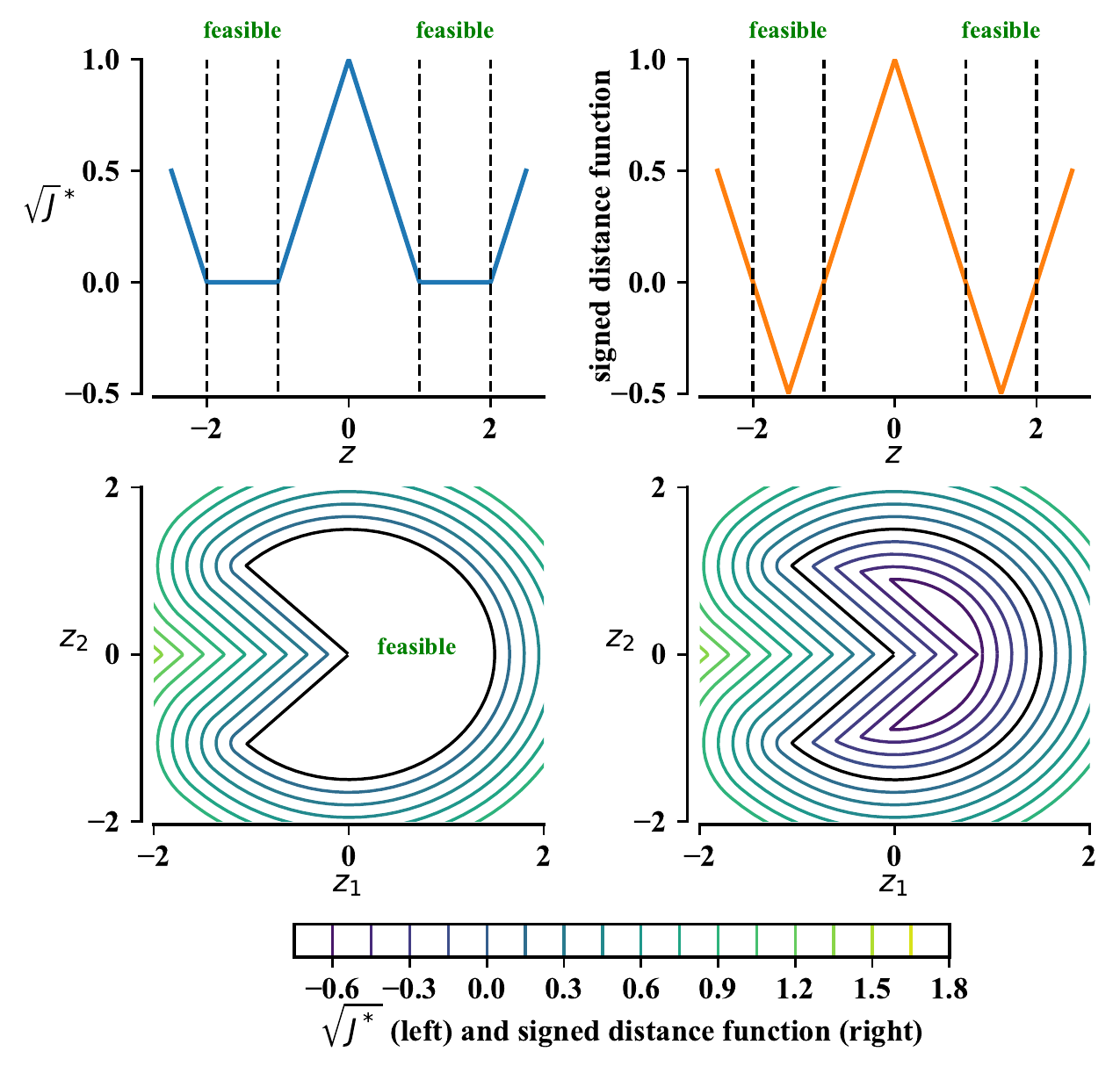}
    \caption{Plots of the square root of the subspace value function $\sqrt{J^*}$ as well as the signed distance function of the boundary for an two imaginary problems. 
    \emph{Top:} The feasible set is the union of the $[-2, -1]$ and $[1,2]$. \emph{Bottom:} The feasible set is 2-dimensional and resembles a Pac-Man character.
    In this project, we have data about $\sqrt{J^*}$, but find it more efficient for classification to aim at representing the signed distance function.
    }
    \label{fig:sdf}
\end{figure}

\subsection{Additional Useful Properties of \texorpdfstring{$\sqrt{J_i^*}$}{Ji}}
\label{sec:sdf}

For any infeasible point $\glb$, $\sqrt{J^*(\glb)}$ is by definition the distance between $\glb$ and the feasible set. Hence, there exist no feasible point in a ball of radius $\sqrt{J^*(\glb)}$ around $\glb$. For the trained neural network $h$ that represents this distance, this property is immediately equivalent to being 1-Lipshitz:
\begin{equation*}
\forall (\glb_1, \glb_2), \quad \Vert h(\glb_1) - h(\glb_2) \Vert \leq \Vert \glb_1 - \glb_2 \Vert
\end{equation*}
Moreover, as can be seen on figure \ref{fig:sdf}, $\sqrt{J^*(\glb)}$ is a continuous function. But its derivative can be non-continuous if the feasible set is not convex, which discourages the use of any continuously differentiable activation function in the neural network $h$.

These properties lead to constraining the type of neural network. In the remaining examples of this paper, Lipshitz networks designate feedforward neural networks with orthogonal weight matrices and GroupSort activation function, as described in \citet{anil2019}. This choice is compatible with fitting non-continuously differentiable functions and allows to automatically enforce 1-Lipshitz continuity.

Finally, any function $h$ that represents a distance to a set respects the following basic properties. For almost every $\glb$ outside of the set:
\begin{eqnarray}
\label{eq:sdf}
    \Vert \nabla h(z) \Vert &=& 1 \\
    h\left(z - h(z)\cdot\nabla h(z)\right) &=& 0 \\
    \nabla{h\left(x - h(z)\cdot\nabla h(z)\right)} &=& \nabla{h(x)}
\end{eqnarray}
This property is enforced through regularization. At each training iteration, samples are randomly drawn within the domain, and a term proportional to the mean squared violation is added to the loss function.
Importantly, we apply this regularization both inside and outside the feasible domain, which encourages $h$ to represent a \emph{Signed Distance Function} (SDF) \citep{osher2003signed} of the feasibility boundary, as shown on the right graphs of figure \ref{fig:sdf}.

\subsection{Performance comparison as dimension increases}

While most fitting approaches do well in low dimension, higher dimensional problems typically reveal more weaknesses. 
Building onto the disk example of figure \ref{fig:relu}, we let the dimension increase while maintaining the same number of points. 
We compare the ability to classify unseen data (as feasible or unfeasible) of four neural network training approaches:

\begin{table}[h!]
 \renewcommand\arraystretch{1.5}
\begin{center}
 \begin{tabular}{|p{1in} p{2.in} p{1.5in} |} 
 \hline
 Name & Loss function & Network type  \\ [0.5ex] 
 \hline
 $J^*$-fit & Mean squared error regression on $J$ & Tanh, feedforward  \\ 
 Classifier & Hinge loss & Tanh, feedforward \\
 Hybrid & Mixed loss (see eqn. \ref{eq:loss}) & Tanh, feedforward \\
SDF & Mixed loss (see eqn. \ref{eq:loss}), with \newline
 SDF regularization (see eqn. \ref{eq:sdf}) & Lipshitz network \\
 \hline
\end{tabular}
\end{center}
\end{table}

\begin{figure}[ht]
    \centering
    \includegraphics[width=0.9\columnwidth]{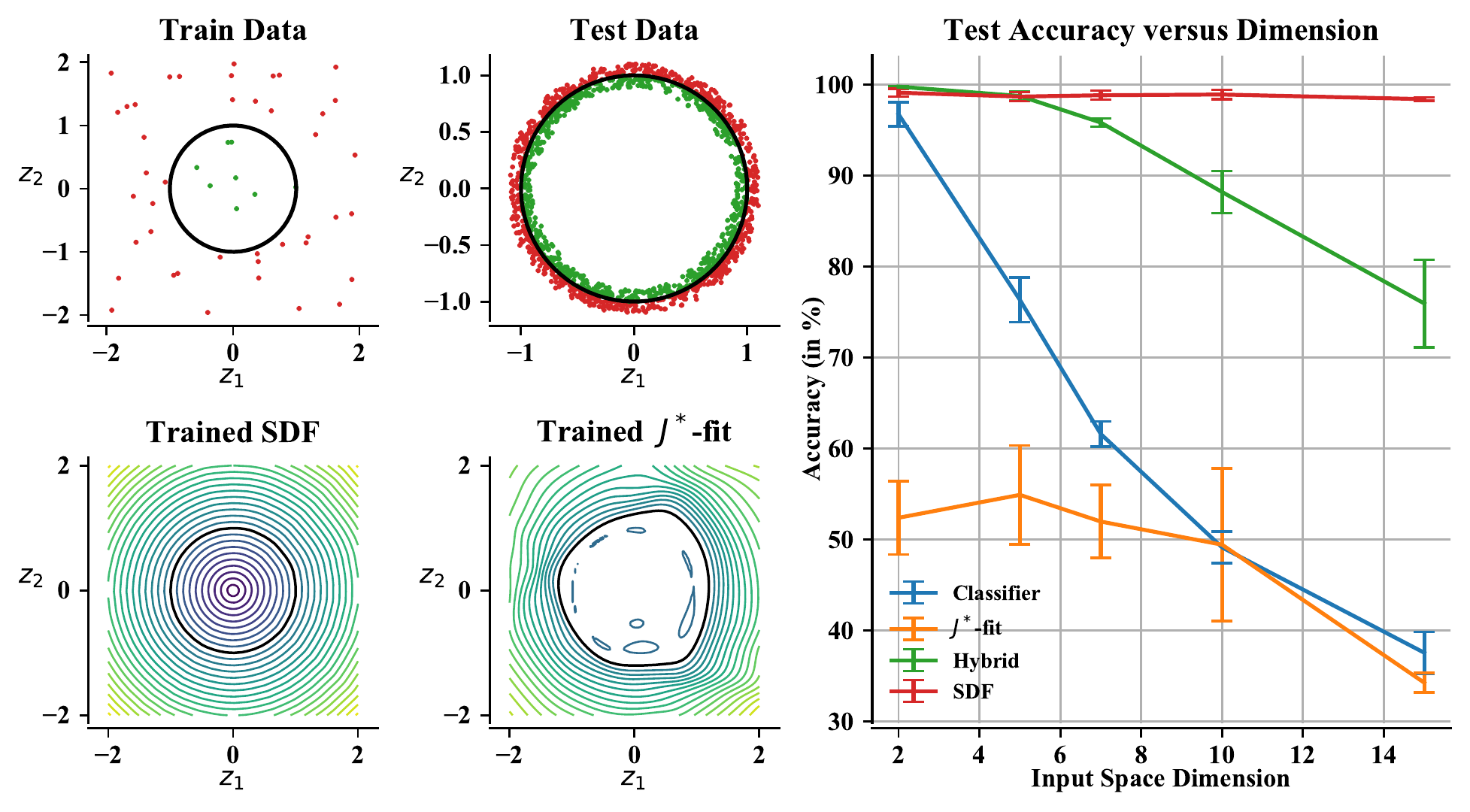}
    \caption{Demonstration of the benefit of fitting a signed distance function to classify disciplinary feasible points given the type of data available in this work. In 2-d, the top left plots show the train and test data while the bottom left ones show the contour lines of the learned network in this case. 
    Fitting $J^*$ directly is harder, and also requires choosing an arbitrary numerical threshold for classification, whereas the SDF only requires a sign check.
    On the right, the input dimension is increased while maintaining the same number of datapoints, demonstrating a much better generalization using the SDF approach.
    The experiment is repeated 5 times.}
    \label{fig:disk_xp}
\end{figure}

The results are shown on figure \ref{fig:disk_xp}, and clearly show that much better generalization is obtained using the signed distance function approach.
The details of the experiment can be found in appendix \ref{appendix:disk_xp}.

\section{Solution to the Aircraft Design Problem}
\label{sec:alg}

The aircraft design problem of section \ref{example} is solved using the neural network classification approach outlined in the previous section. We adopt a very simple design of experiment: 
\begin{enumerate}
    \item Sample points  $\left(\glb^{(j)}\right)_{j \in 1 \dots N_{ini}}$ randomly in the domain and evaluate $J^*_i(\glb^{(j)})$ for each discipline $i$.
    \item Repeat: 
    \begin{enumerate}
        \item For each discipline $i$, fit a SDF neural network $h_i$ following the methods of section \ref{sec:sdf}.
        \item Find a new candidate point $\glb$ and add it to the dataset:
         \begin{equation}
         \label{eq:candidate}
         \glb = \argmin_{\glb \in \{\glb \vert \forall i \in 1 \dots N_d \; h_i(\glb) \leq 0\}}  f (\glb)
         \end{equation}
    \end{enumerate}
\end{enumerate}
If there is no solution to equation \ref{eq:candidate}, we use the point that is closest to being feasible.
While our work focuses on surrogate modeling approaches, more principled exploration could improve the performance on more complex problems.

Table \ref{table:alg} compares the performance of our approach with a direct application of CO where the system level problem is solved with sequential quadratic programming \citep{Braun1996}, as well a with a conventional surrogate modeling approach using Gaussian process (GP) regression on $J$. Appendix \ref{appendix:final_xp} gives details about the algorithm and the baselines, and shows visualizations of the SDF network's build up.

\begin{table}[h!]
\label{table:alg}
 \renewcommand\arraystretch{1.5}
\begin{center}
 \begin{tabular}{|p{1.1in}p{1in} p{1.2in} p{1.in}|} 
 \hline
  & Conventional CO with SQP & Gaussian Process surrogate model of $J^i$  & Signed Distance approach (ours)\\ 
 \hline
 No. of system-level iterations & 74 (+/- 28)&  18 (+/- 6)& 6 (+/- 2) \\
  No. of aerodynamic function evaluations & 2161 (+/- 796)&  267 (+/- 63) & 110 (+/- 57) \\
  No. of propulsion function evaluations& 496 (+/- 209) &  94 (+/- 23)& 31 (+/- 8) \\
 \hline
\end{tabular}
\end{center}
\caption{Solution of the aircraft design problem of section \ref{example} using previous CO approaches and with our approach. The table shows the number of iterations to 5\% of the optimal value. The rounded mean and standard deviation over 20 trials are indicated. The two last column clearly demonstrate the advantage of using a surrogate model over the original CO architecture. The performance of our approach comes from the resolution of the ambiguous decision rule described in section \ref{sec:mix}, as well from a family of function adequate to this particular problem.}
\end{table}

\section{Conclusion and Future Work}
\label{sec:disc}
The convergence speed of CO, a distributed multidisciplinary design optimization architecture is improved by using neural networks to classify target shared variables as feasible or infeasible. For infeasible points, the training data contains the distance to the closest feasible point and its derivative. Leveraging this information by training Lipshitz networks with a custom asymmetric loss function and proper regularization, we show that more reliable and faster convergence can be obtained.


An important drawback of the approach presented here is the lack of a principled exploration strategy. 
In typical Bayesian optimization \citep{Frazier2018, Snoek2013, Snoek2015, Hernandez-Lobato2016}, the exploration-exploitation trade off is carefully taken care of by optimizing an acquisition function such as expected improvement.
This limitation will be addressed in future work, for instance by building upon the work in \citep{Snoek2015}.

Future work will also consist in applying our approach to more complex and larger scale problems.  Further extensions could develop multi-fidelity and multi-objective versions \citep{multiobjco2009, m2002multi}, for which surrogate modeling and Bayesian optimization are already often used \citep{picheny2015multiobjective, Rajnarayan2008, Meliani2019}.



\paragraph{Acknowledgment:} This research is funded by the King Abdulaziz City for Science and Technology through the Center of Excellence for Aeronautics and Astronautics. \url{https://ceaa.kacst.edu.sa/}

\bibliographystyle{unsrtnat}
\bibliography{mdorefs,bayesopt}

\appendix

\section{Details of the Aircraft Design Example Problem}
\label{appendix:example}

\paragraph{Nomenclature}
$\cpyVar X$ is the subspace-level local copy of the system-level global variable $\glbVar X$.
\begin{longtable}{ll}
$D_{wing}, D_{fixed}, D$ & Wing drag, fixed drag (drag of everything but the wing), total drag \\
$\eta$ & Propulsion system efficiency \\
$V$ & Cruise speed \\
$L$ & Lift \\
$b, S$ & Wing span and wing area \\
$W_{wing}, W_{bat}, W_{fixed}, W$ & Wing weight, battery weight, fixed weight (everything but the wing), total weight\\
$\omega$ & propeller RPM \\
$T$ & Thrust \\
$Q_p, Q_m$ & Propeller and motor torque magnitude \\
$U$ & Motor voltage \\
$P_{in}$ & Electric power consumed by the motor \\
$l_r$ & Desired Range \\
\end{longtable}

\paragraph{Propulsion System Model:} The motor voltage $U$ and propeller RPM $\omega$ are the design variables for this discipline. U is bounded between 0 and 9V, and the $\omega$ between 5000 and 10000.

The motor model is simply a 3 constants model (see for instance \citep{Drela2007}). We use a Turnigy D2836/9 950KV Brushless Outrunner (Voltage 7.4V - 14.8V, Max current: 23.2A).
\begin{enumerate}
    \item $K_v$ = 950 RPM/V
    \item $R_M$ = 0.07 $\Omega$
    \item $I_0$ = 1.0 A
\end{enumerate}
We use the following model to compute the motor torque and  as a function of the voltage:
\begin{align*}
    I &= (U - \omega / K_v) / R \\
    Q_m &= \left(I - I_0\right)\frac 1 K_q
\end{align*}
Here we choose $K_q$ as equal to $K_v$ (converted to units of A/Nm).

The propeller model uses measurements from the UIUC propeller database \citep{UIUC2020} for a Graupner 9x5 Slim CAM propeller data.
The propeller radius is $R = 0.1143 m$.
The torque and thrust data for the propeller are fitted using polynomials:
\begin{align*}
    Q_p  &= \rho  4/\pi^3 R^3  (0.0363\omega^2 R^2 + 0.0147 V \omega R \pi - 0.0953 V^2 \pi^2) \\
    C_T &= 0.090 - 0.0735J - 0.1141J^2 \\ 
    T &= C_T * \rho * n^2 * (2*R)^4; \\
    n &= \omega / (2\pi) \\
    J &= V / n / R 
\end{align*}

Finally the battery weight is computed using a battery energy density of $\nu_e=720e3$ J/kg.
The power consumed by the propulsive system is  $P_{in} = IU$, and
the flighttime for a distance $l_r = 42000$m is $t=l_r/V$, so we get:
\begin{equation*}
    W_{bat} = \frac{P_int}{\nu_e}=\frac{IUl_r}{V \nu_e}
\end{equation*}

\paragraph{Wing Model:} The wing local variables are the span $b$, the wing area $S$, the lift $L$ and the wing weight $W_{wing}$. 

Geometrically the wing is assumed to have a taper ratio $t_r = 0.75$, a form factor $k=2.04$, a thickness to chord ratio $t_c=0.12$ and a tail area ratio $t_{t}=1.3$. The area ratio of the airfoil is $k_{airfoil}=0.44$. The fixed weight of the fuselage, motor and propeller is $W_fixed=37.28N$. The fuselage has a wetted area $S_f = 0.18m^2$, a body form factor $k_f=1.22$ and a length of $l_f = 0.6m$. We use $g = 9.81 N/kg$ and $\rho=1.225kg/m^3$. 

We first use simple aerodynamic theory to compute the lift and the drag of the wing. We assume $C_{L, max}=1.$ and a span efficiency factor $e=0.8$. The dynamic viscosity of the air on that day is chosen to be $\nu=1.46e-5m^2/s$.
The lift simply is equal to the weight:
\begin{align*}
    W &= W_{fixes} + W_{bat} + W_{wing} \\
    L &= W \\
\end{align*} 
Then, the induced drag coefficient can be computed:
\begin{align*}
    C_L &= 2 L/ (S \rho V^2)  \\
    C_{Di} &= C_L^2 S/(\pi b^2e) \\
\end{align*}
The parasitic drag is computed assuming fully turbulent flow on the wing and the fuselage:
\begin{align*}
    Re_{wing} &= \frac{V S}{b\nu} \\
    C_{f,wing} &= 0.074/Re_{wing}^{0.2} \\
    C_{Dp,wing} &= (1+2t_c)C_{f,wing}kt_t \\
    Re_{fuse} &= \frac{V l_f}{\nu} \\
    C_{f,fuse} &= 0.074/Re_{fuse}^{0.2} \\
    C_{Dp,fuse} &= C_{f,fuse} \frac {S_f} S k
\end{align*}
Finally we add a penalty $C_{Ds}$ for post-stall flight and complete the final drag:
\begin{align*}
    C_{Ds} &= 0.1 \max(0, C_L - C_{L, max})^2 \\
    D &= \frac 1 2 \rho V^2 S \left( C_{Di} +C_{Dp,wing} + C_{Dp,fuse} + C_{Ds}\right)
\end{align*}

The wing is assumed to be made of a main carbon spar along with styrofoam. The foam weight is simply related to the wing area. We use $\rho_{foam} = 40 kg/m^3$
\begin{equation*}
    W_{foam} = g \rho_{foam} \frac {S^2}{b} t_c k_{airfoil} \\
\end{equation*}

The gauge of the carbon spar is computed based on stress and deflection, with a maximum thickness of $\tau_{m} =1.14mm$.
The maximum possible stress is $\sigma_{max} = 4.413e9 N/m^2$, the Young modulus is $E=2.344e11 N/m^2$ and the density is $\rho_{carbon} = 1380 kg/m^3$.
The radius of the spar is computed using the wing dimensions: $r_s = \frac S {4b}  t_c$.
The minimum carbon thickness to withstand the stress is computed by:
\begin{align*}
        M_{root} &= L b/8 \\
        I &= \pi r_s^3 \tau_m \\
        \tau_{stress} &= \tau_{m}Lb/8 r_s /I / \sigma_{max}/0.07
\end{align*}
The minimum carbon thickness to avoid deflections is computed by:
\begin{align*}
         \delta = L  b^4 / (64EI) \\
        \tau_{defl} = \tau_{m} \frac{2\delta}{b} /0.07
\end{align*}
The actual gauge is the thickest one: $\tau = \max(\tau_{defl}, \tau_{stress}, \tau_m)$.
The total mass can then be computed:
\begin{align*}
        W_{spar} &= 2 \pi r_s \tau b \rho_{carbon} \\
        W_{wing} &= W_{spar} +W_{foam} 
\end{align*}

The optimal value of the problem happens for 
\begin{longtable}{ll}
D &= 2.15$N$ \\
V &= 13.71 $m/s$\\
b &= 2.7574 $m$ \\
S &= 0.397 $m^2$\\
RPM &= 8127 \\
U &= 9$V$
\end{longtable}

\section{Including the Subspace Problem Meta-Information in the Loss Function}
\label{appendix:subspace}

As mentioned in the main paper, the subspace optimization problem is a projection onto the disciplinary feasible set.

\begin{equation}
    \label{eq:sub_}
    J_i^*(\glb) = \min_{\cpy \in \{ \cpy \vert \exists \loc \; c_i(\loc, \cpy) \leq 0\}} \Vert \cpy - \glb \Vert_2^2
\end{equation}

\citep{Braun1996} proves that, for each evaluation $J_i^*(\glb)$, we can compute the gradient of the converged point:
\[\nabla J_i(\glb) = (\glb - \cpy^*)\]
where $\cpy^*$ is the optimal point of the subspace evaluation \ref{eq:sub_}.
We also know from \ref{eq:sdf} that $J^i(\cpy^*) = 0$, and $\nabla J^i(\cpy^*) = \nabla J^i(\glb)$,
In cases like section \ref{sec:regcls} where we are interested in information about $\sqrt{J^*_i}$, the gradient information can be derived easily from this using the chain rule.

A slightly modified version of the loss function of equation \ref{eq:loss} is then:
\begin{equation}
    \label{eq:loss_}
    l(\glb^{(j)}) = \begin{cases}
                    \begin{array}{cc}
                     \vert h_i(\glb^{(j)}) - \sqrt{{J_i^*}^{(j)}} \vert  &+
                     \Vert \nabla h_i(\glb^{(j)}) -  \nabla \sqrt{J_i^*}^{(j)} \Vert_1 +\\
                    \vert h_i({(\cpy^*)}^{(j)}) \vert  &+
                    \Vert \nabla h_i({(\cpy^*)}^{(j)}) - \nabla \sqrt{J_i^*}^{(j)} \Vert_1 
                    \end{array}
                    & \text{if } \glb^{(j)} \text{ is infeasible}
                
                \\        
                
                \max(h_i(\glb^{(j)}), 0) & \text{otherwise}
                \end{cases}
\end{equation}

\section{Details on the Fits of a Disk of Increasing Dimension}
\label{appendix:disk_xp}

In this experiment, there are 5 different training sets each containing 50 points, while the test set remains the same and contains 500 points. The experiment is repeated on each train sets in order to obtain error bars on the reported performance.

All feedforward networks have 3 layers of 8 units each. They are trained with the Adam Optimizer with $ \beta=(0.9, 0.999)$ and a $10^{-3}$ learning rate tuned with gridsearch.

The Lipshitz networks use linear layers orthonormalized after each update using 15 iterations of Bjorck's procedure of order 1 with $\beta = 0.5$.
They use a full vector sort as activation function (GroupSort with groupsize 1).

\section{Details on the Solution to the Aircraft Design Problem}
\label{appendix:final_xp}

\subsection{Details and Parameter Choices}

The system level problem is solved 20 times using each approach, using the same random seed for each solution approach.
The wing and propulsion subspace problems (see equation \ref{eq:sub}) are solved using the sequential quadratic programming solver SNOPT \citep{gill2005snopt}. 

The airspeed $V$ is bound to be between 5 and 15 $m/s$, the battery mass between $0.1$ and $1$ $kg$, and the drag $D$ between 1 and 6 $N$.

For both surrogate model approaches (Gaussian process and SDF), the initial number of samples $N_{INI}$ is chosen to be simply 1, which allows to compare with the direct approach more easily.
The choice of a new candidate point is then done by optimizing the surrogate model, as shown in \ref{eq:candidate}. 
This optimization is also performed by SNOPT, with 15 random restarts.

\paragraph{Regular CO:} For regular CO, the system level problem is solved directly using SNOPT. A feasibility threshold of $10^{-4}$ is used to help convergence.

\paragraph{Gaussian processes baseline:} At each iteration, a Gaussian Process is fit to the currently available data for $J_i^*$, as well as the gradient information. The prior mean is simply the zero function, and the kernel is a scaled squared-exponential kernel. The hyper parameters are tuned using type II maximum likelihood minimization and the Adam Optimizer with $ \beta=(0.9, 0.999)$ and a $10^{-3}$ learning rate for 200 iterations. The code uses the GpyTorch framework \citep{gardner2018gpytorch}.
Again, a feasibility threshold of $10^{-4}$ is required to help convergence.

\paragraph{SDF neural networks:} The SDF surrogate models use three layers of 12 units with GroupSort activation functions (groupsize of 1). The  linear layers are orthonormalized after each update using 15 iterations of Bjorck's procedure with order 1 and $\beta = 0.5$.  They are trained with the Adam optimizer with $ \beta=(0.9, 0.999)$ and a learning rate that is tuned at each iteration (we use the best network after out of 10 randomly sampled learning rates between $10^{-5}$ and $10^{-3}$).

\end{document}